\documentclass{article}
\usepackage{ijcai24}

\usepackage{times}
\usepackage{soul}
\usepackage{url}
\usepackage[hidelinks]{hyperref}
\usepackage[utf8]{inputenc}
\usepackage[small]{caption}
\usepackage{graphicx}
\usepackage{amsmath}
\usepackage{amsfonts}
\usepackage{amsthm}
\usepackage{booktabs}
\usepackage{algorithm}
\usepackage{algorithmic}
\usepackage{multirow}
\usepackage[switch]{lineno}
\usepackage{pifont}
\usepackage{makecell}
\usepackage{hyperref}
\makeatletter
\def\UrlAlphabet{%
      \do\a\do\b\do\c\do\d\do\e\do\f\do\g\do\h\do\i\do\j%
      \do\k\do\l\do\m\do\n\do\o\do\p\do\q\do\r\do\s\do\t%
      \do\u\do\v\do\w\do\x\do\y\do\z\do\A\do\B\do\C\do\D%
      \do\E\do\F\do\G\do\H\do\I\do\J\do\K\do\L\do\M\do\N%
      \do\O\do\P\do\Q\do\R\do\S\do\T\do\U\do\V\do\W\do\X%
      \do\Y\do\Z}
\def\UrlDigits{\do\1\do\2\do\3\do\4\do\5\do\6\do\7\do\8\do\9\do\0}
\g@addto@macro{\UrlBreaks}{\UrlOrds}
\g@addto@macro{\UrlBreaks}{\UrlAlphabet}
\g@addto@macro{\UrlBreaks}{\UrlDigits}
\makeatother

\title{Fine-Grained Zero-Shot Learning: Advances, Challenges, and Prospects}

\author{
Jingcai Guo$^{1,2}$
\and
Zhijie Rao$^{1}$
\and
Zhi Chen$^{3}$
\and
Jingren Zhou$^{4}$\And
Dacheng Tao$^{5}$\\
\affiliations
$^1$The Hong Kong Polytechnic University, Hong Kong SAR\\
$^2$Hong Kong Polytechnic University Shenzhen Research Institute, China\\
$^3$The University of Queensland, Australia\\
$^4$Alibaba Group, China\\
$^5$The University of Sydney, Australia\\
\emails
\{jc-jingcai.guo, zhijie.rao\}@polyu.edu.hk, 
zhi.chen@uq.edu.au,\\
jingren.zhou@alibaba-inc.com, 
dacheng.tao@sydney.edu.au
}

\begin{document}

\maketitle

\begin{abstract}
%
%
Recent zero-shot learning (ZSL) approaches have integrated fine-grained analysis, i.e.,~\textit{fine-grained ZSL}, to mitigate the commonly known seen/unseen domain bias and misaligned visual-semantics mapping problems, and have made profound progress. 
Notably, this paradigm differs from existing close-set fine-grained methods and, therefore, can pose unique and nontrivial challenges. 
However, to the best of our knowledge, there remains a lack of systematic summaries of this topic. 
To enrich the literature of this domain and provide a sound basis for its future development, in this paper, we present a broad review of recent advances for fine-grained analysis in ZSL. 
Concretely, we first provide a \textit{taxonomy} of existing methods and techniques with a thorough analysis of each category. Then, we summarize the \textit{benchmark}, covering publicly available datasets, models, implementations, and some more details as a library\footnote{Accessible via \url{https://github.com/eigenailab/Awesome-Fine-Grained-Zero-Shot-Learning}}. Last, we sketch out some related \textit{applications}. In addition, we discuss vital \textit{challenges} and suggest \textit{potential future directions}.
\end{abstract}

\section{Introduction}
Conventional recognition tasks are mostly performed in a \textit{close-set} scenario, i.e., the test categories are subsets or, at most, identical to the training categories. However, such close-set models may fail in real-world applications where novel categories can easily appear.  
With the goal of extending recognition to unseen categories, zero-shot learning (ZSL)~\cite{lampert2009learning_cvpr09} has emerged and attracted lots of interest in the machine learning and computer vision communities. 
Practically, ZSL can be formulated as a \textit{visual-to-semantics} mapping problem by using a set of semantic descriptors shared by both seen and unseen categories. Such semantics are high-level, per-category, and more importantly, much more accessible than labeled real data samples, such as word~\cite{welinder2010caltech} or sentence~\cite{nilsback2008automated} descriptions as the bridge for knowledge transfer.

\begin{figure}[t]
    \centering
    \begin{minipage}{\linewidth}
    \centerline{\includegraphics[width=0.81\textwidth]{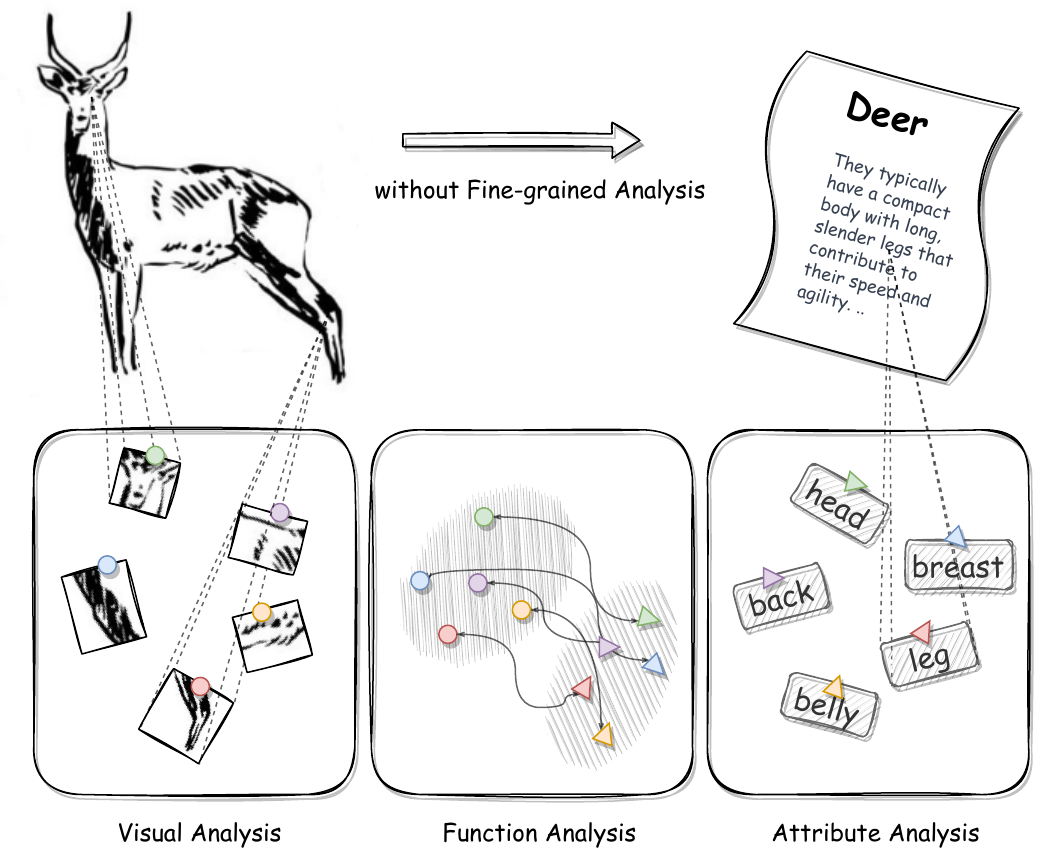}}
    \end{minipage}
    \caption{Compared with conventional ZSL, which generally studies class-wise relations, FZSL incorporates more refined and delicate concepts typically embodied in three realms of analysis, including \textbf{\textit{Visual}}, \textbf{\textit{Attribute}}, and \textbf{\textit{Mapping Function}}.}
    \label{fig:intro}
\vspace{-4mm}
\end{figure}

Since there is no observation of any unseen category samples, the trained models are inherently biased to seen categories, i.e., \textit{domain bias}~\cite{fu2015transductive_pami15}. Moreover, 
the visual features and semantics are also mutually independent, thus further challenging their \textit{alignment}~\cite{li2023vs_ijcai23}. 
Traversing the literature, most ZSL methods approach the visual-to-semantics problem by extracting each sample's global features in a coarse-grained manner. However, it inevitably degrades the overall recognition, especially for those samples with small inter- and large intra-variation between categories, 
%
e.g., the visual differences between various `\textit{husky subspecies}' can be far greater than the differences between `\textit{husky}' and `\textit{wolf}'. 
%
To better mitigate these problems, recent ZSL studies have focused increasingly on the fine-grained aspects and obtained huge progress in terms of theories, algorithms, and applications~\cite{ji2018stacked_nips18,huynh2020fine_cvpr20,guo2023graph}.

Observations reveal that fine-grained ZSL~(FZSL) is more favorable to transferring knowledge between seen/unseen categories, wherein its gist is to capture subtle visual differences that are not only discriminative between categories, but also well-aligned to their diverse and complex semantics. 
Despite recent progress in FZSL, a thorough overview summarizing its advances, challenges, and prospects is not available yet. To fill the gap, this paper aims to systematically review the current development of FZSL, covering a wide range of methods and techniques used in the fine-grained extension of ZSL, and further provide a basis for its future development. 
In a nutshell, our contributions are four-fold, i.e., 
\begin{itemize}
    \item We propose a comprehensive taxonomy of FZSL and provide a thorough analysis of the methods and techniques behind it~(\textbf{Section~\ref{taxonomy}}), which assists researchers with a better exploration of their interests.
    %
    \item We provide a library to facilitate an overview of commonly used datasets, specific experimental setups, and other details~(\textbf{Section~\ref{library}}).
    \item We sketch out a series of the most representative FZSL applications in various domains~(\textbf{Section~\ref{application}}), which initiates interdisciplinary research and vision.
    \item We discuss vital challenges in this domain and share our insights on the future research direction~(\textbf{Section~\ref{c_and_o}}), which concludes this first survey on FZSL.
\end{itemize}

\section{Problem Formulation}


%
Given the seen domain $\mathcal{D}^s=\{(x^s, y^s, a^s)|x^s\in\mathcal{X}^s, y^s\in\mathcal{Y}^s, a^s\in\mathcal{A}^s\}$, where $\mathcal{X}^s$, $\mathcal{Y}^s$, and $\mathcal{A}^s$ denote visual samples, category labels, and semantics (e.g., a set of attributes), 
%
and similarly, let $\mathcal{D}^u=\{(x^u, y^u, a^u)|x^u\in\mathcal{X}^u, y^u\in\mathcal{Y}^u, a^u\in\mathcal{A}^u\}$ denote the unseen domain. 
Without loss of generality, the task of ZSL can be modeled as learning a mapping/relational function $\Psi: \mathcal{X}^s \to \mathcal{A}^s$, wherein $\mathcal{X}^u$ is strictly inaccessible for training. 
During inference, the learned function $\Psi$ is applied to recognize samples from the unseen domain only, i.e., ZSL, or from the joint of both seen and unseen domains, i.e., Generalized ZSL (GZSL)\footnote{For simplicity, we use ZSL to refer to both ZSL and GZSL scenarios in the remaining sections of this survey.}. 
Notably, the success of ZSL relies on the sharing property between $\mathcal{A}^s$ and $\mathcal{A}^u$, which act as the bridge from seen to unseen domains.

Category-wise relational modeling has achieved promising results as the most common practice to approach the ZSL problem, with the recognition objective as:
\begin{equation}
    \mathop{\arg \min}_{\Psi} \  P(y|\Psi(x, a)),
\end{equation}
where $P$ is the posterior probability and $\Psi$ denotes the relational function. However, class-wise modeling exhibits unavoidable limitations on fine-grained recognition tasks due to the erasure of large amounts of information. 
In recent years, extensive studies have embedded fine-grained analysis into ZSL to achieve a more refined modeling capability, i.e., fine-grained ZSL (FZSL) as shown in Figure \ref{fig:intro}, with a derivative recognition objective as:
\begin{equation}
\mathop{\arg \min}_{\Psi, \Phi, \Theta} \  P(y|\Psi(\Phi(x), \Theta(a))),
\end{equation}
where $\Phi$, $\Theta$, and $\Psi$ represent fine-grained \textbf{Visual}, \textbf{Attribute}, and \textbf{Function} analysis, respectively. 
In this paper, we summarize the efforts of research for the FZSL community over the last few years, which have driven one or more remarkable advances in the aspects of $\Phi$, $\Theta$, and $\Psi$.

\section{Taxonomy}
\label{taxonomy}

\subsection{Overview}
We empirically categorize FZSL models into two broad directions: \textbf{Attention-Based} methods~(elaborated in Table~\ref{tab:attention}) and \textbf{Non-Attention} methods~(elaborated in Table~\ref{tab:non-attention}). 
Concretely, attention-based methods follow the most intuitive motivation of shifting the global view to multiple local views to focus on the most valuable parts. In this direction, we further categorize representative studies into three primary areas, including \textit{Attribute Attention}, \textit{Visual Attention}, and \textit{Cross Attention}, according to the targets on which the attention mechanisms act, and further tag secondary areas for them in terms of concrete implementations. 
Meanwhile, for the direction of non-attention methods, we categorize them according to their core motivation as well as specific designs, including \textit{Prototype Learning}, \textit{Data Manipulation}, \textit{Graph Modeling}, \textit{Generative Method}, and \textit{Others} as the primary areas. 
It is important to note that some methods can cover more than one area, and we categorize them according to their most critical module. 

\subsection{Preliminaries}
We elaborate on some of the basic elements and terminologies in Table~\ref{tab:attention} and Table~\ref{tab:non-attention}. 
%
\textbf{Attribute-Free} indicates that no fine-grained attribute annotations are required, which can refer to professional-level annotations, e.g., describing a deer by using detailed information of \textit{\{head, breast, leg, etc.\}}. Attribute-free methods usually require only class-wise semantic embeddings or even no semantic guidance. Note that we only discuss whether the core component of a method is attribute-free or not, not for its entire framework. 
\textbf{Auxiliary} denotes the auxiliary information used in addition to attribute annotations. For example, some methods resort to external resources to gain additional prior knowledge~\cite{liu2021goal} or to release the restriction of fine-grained attribute annotations~\cite{elhoseiny2017link}. Some typical information includes \textit{Gaze Annot}, i.e., human visual attention annotation; \textit{Region Annot}, i.e., local visual annotation; and \textit{Online Media}, i.e., the language library for obtaining attribute descriptions~\cite{naeem2022i2dformer}. 
%

\subsection{Attention-Based Methods}

\begin{table*}[t]
    \renewcommand{\arraystretch}{0.77}
    \centering
    \resizebox{\textwidth}{!}{
    \begin{tabular}{l l l c c}
        \toprule
        \textbf{\large{Primary Area}} & \textbf{\large{Secondary Area}} & \textbf{\large{Method}} & \textbf{\large{Attribute-Free}} & \textbf{\large{Auxiliary}}\\
        \midrule
        \textit{Attribute Attention} & \textit{Normalized Weight} & LFGAA \cite{liu2019attribute} & \ding{55} & \ding{55}\\
        \midrule
        \multirow{7}{*}{\textit{Visual Attention}} & \textit{Normalized Weight} & LAPE \cite{wang2022language} & \ding{55} & \ding{55}\\ 
        \cmidrule{2-5}
        & \multirow{4}{*}{\textit{Attention Mask}} & AREN \cite{xie2019attentive} & \ding{51} & \ding{55}\\
         &  & RGEN \cite{xie2020region} & \ding{51} & \ding{55}\\
         &  & RSAN \cite{wang2021region} & \ding{55} & \ding{55}\\
         \cmidrule{2-5}
         & \multirow{2}{*}{\textit{Local Coordination}} & LDF \cite{li2018discriminative} & \ding{51} & \ding{55}\\
         &  & SGMA \cite{zhu2019semantic} & \ding{51} & \ding{55}\\
        \cmidrule{1-5}
        \multirow{8}{*}{\textit{Cross Attention}} & \multirow{3}{*}{\textit{Score Function}} & DAZLE \cite{huynh2020fine_cvpr20} & \ding{55} & \ding{55}\\
         &  & GEM \cite{liu2021goal} & \ding{55} & \textit{Gaze Annot}\\
         &  & MSDN \cite{chen2022msdn} & \ding{55} & \ding{55}\\
         \cmidrule{2-5}
         & \multirow{5}{*}{\textit{Self Attention}} & TransZero \cite{chen2022transzero} & \ding{55} & \ding{55}\\
         &  & I2DFormer \cite{naeem2022i2dformer} & \ding{51} & \textit{Online Media}\\
         &  & DUET \cite{chen2023duet} & \ding{55} & \ding{55}\\
         &  & PSVMA \cite{liu2023progressive} & \ding{55} & \ding{55}\\
         &  & HRT \cite{cheng2023hybrid} & \ding{55} & \ding{55}\\
        \bottomrule
    \end{tabular}}
    \centering
    \caption{The categorization of representative attention-based fine-grained zero-shot learning methods.}
    \label{tab:attention}
\end{table*}

As shown in Table~\ref{tab:attention}, attention-based methods are the most intuitive and natural primary areas for FZSL. 
Among them, \textit{Attribute Attention} and \textit{Visual Attention} aim at focusing on the most valuable subattributes and local visual regions/parts, respectively. In contrast, \textit{Cross Attention} seeks to capture correlation links between local visual regions and subattributes. Further, we categorize them more in-depth according to their specific implementation strategies of the attention mechanism, including \textit{Normalized Weight}, \textit{Attention Mask}, \textit{Local Coordination}, \textit{Score Function}, and \textit{Self-Attention}.

\subsubsection{Normalized Weight}
The motivation of normalized weight is to learn a one-dimensional vector for weighting attentional targets, thus suppressing the influence of extraneous regions/parts. 
Among them, LFGAA \cite{liu2019attribute} applies it for attribute attention inspired by the observation that different attributes are not equally important for sample category determination. The gist of such methods is to adaptively filter the most significant attributes based on visual features, whose formula can be expressed as:
\begin{equation}
    W_a = \frac{\exp(\mathcal{F}(x))}{\sum^{m}\exp(\mathcal{F}(x))},
\end{equation}
where $W_a\in\mathbb{R}^{m}$ is the normalized weight and $m$ denotes the dimension of attribute. $\mathcal{F}$ denotes the learnable network, and $x$ is the visual feature. Then, it multiplies the weight vector with the attribute vector to suppress unimportant attributes. 

In contrast, LPAE~\cite{wang2022language} applies normalized weight to visual attention. Specifically, suppose that $x\in \mathbb{R}^{C\times H\times W}$ denotes the visual feature of a sample with $r=H\times W$ regions, where $C$, $H$, and $W$ are the dimension, height, and weight, respectively, and suppose different regions have different importance for category judgment. Therefore, LPAE resorts to learning the weights of regions based on attribute prompts, which can be expressed as:
\begin{equation}
    W_v = \frac{\exp(\mathcal{F}(x, a))}{\sum^{r}\exp(\mathcal{F}(x, a))},
\end{equation}
where $W_v\in \mathbb{R}^{r}$ is the weights, $a$ denotes the attribute vector, and $\mathcal{F}$ is the learnable network. It adopts the idea of self-attention (described later) to design $\mathcal{F}$. After obtaining the normalized weights, it further multiplies the weights with the original features to obtain the enhanced features, which are fed into the downstream network for classification.

\subsubsection{Attention Mask}
The gist of the attention mask is to encourage the learned models to focus on multiple regional visual features simultaneously. Typically, a generative network is usually deployed to generate $N$ masks with the same dimensions as the input features, where each mask reveals a key regional feature. 
It can be expressed as $M=\mathcal{F}(x)$, where $x\in \mathbb{R}^{C\times H \times W}$ is the visual feature, $\mathcal{F}$ denotes the generative network, and $M\in \mathbb{R}^{N \times H \times W}$ denotes $N$ attention masks. 
Afterward, multiplying the masks with the original features yields $N$ regional features, which can be expressed as:
\begin{equation}
    x_{region} = \{xm_1, xm_2, ..., xm_N\},
\end{equation}
where $[m_1, m_2, ..., m_N]=M, m_i \in \mathbb{R}^{H \times W}$. 

The difference between various attention mask methods lies in the way the subsequent processing of $x_{region}$ is carried out. 
For example, AREN~\cite{xie2019attentive} 
employs adaptive thresholding to further filter out the noisy regions/parts and thus assist the classifier in determination. RSAN~\cite{wang2021region} instead uses max-pooling to obtain a one-dimensional vector, which is then aligned with the attribute vector. In contrast, RGEN~\cite{xie2020region} introduces the graph to model the topological relationships between different regions/parts.

\begin{table*}[t]
    \renewcommand{\arraystretch}{1.0}
    \centering
    \resizebox{\textwidth}{!}{
    \begin{tabular}{l l l c c}
        \toprule
        \textbf{\large{Primary Area}} & \textbf{\large{Secondary Area}} & \textbf{\large{Method}} & \textbf{\large{Attribute-Free}} & \textbf{\large{Auxiliary}}\\
        \midrule
        \multirow{6}{*}{\textit{Prototype Learning}} & \multirow{3}{*}{\textit{Prototype-Independent}} & 
        APN \cite{xu2020attribute} & \ding{55} & \ding{55}\\
         &  & CC-ZSL \cite{cheng2023discriminative} & \ding{55} & \ding{55}\\
         &  & CoAR-ZSL \cite{du2023boosting} & \ding{55} & \ding{55}\\
         \cmidrule{2-5}
         & \multirow{3}{*}{\textit{Prototype-Symbiotic}} & DPPN \cite{wang2021dual} & \ding{55} & \ding{55}\\
         &  & DPDN \cite{ge2022dual} & \ding{55} & \ding{55}\\
         &  & GIRL \cite{guo2023group} & \ding{55} & \ding{55}\\
        \midrule
        \multirow{5}{*}{\textit{Data Manipulation}} & \textit{Patch Clustering} & VGSE-SMO \cite{xu2022vgse} & \ding{51} & \ding{55}\\
        \cmidrule{2-5}
        & \multirow{2}{*}{\textit{Detector-Based}} & LH2B \cite{elhoseiny2017link} & \ding{51} & \textit{Region Annot}\\
         &  & S2GA \cite{ji2018stacked_nips18} & \ding{51} & \textit{Region Annot}\\ 
         \cmidrule{2-5}
         & \multirow{2}{*}{\textit{Image Crop}} & SR2E \cite{ge2021semantic} & \ding{51} & \ding{55}\\
         &  & ERPCNet \cite{li2022entropy} & \ding{51} & \ding{55}\\
        \midrule
        \multirow{6}{*}{\textit{Graph Modeling}} & \multirow{4}{*}{\textit{Visual Enhancement}} & RIAE \cite{hu2022region} & \ding{55} & \ding{55}\\
         &  & GNDAN \cite{chen2022gndan} & \ding{55} & \ding{55}\\
        &  & GKU \cite{guo2023graph} & \ding{51} & \textit{Region Annot}\\
         \cmidrule{2-5}
         & \textit{Attribute Enhancement} & APNet \cite{liu2020attribute} & \ding{51} & \ding{55}\\
         \cmidrule{2-5}
         & \textit{Region Search} & EOPA \cite{chen2023explanatory} & \ding{55} & \ding{55}\\
        \midrule
        \multirow{3}{*}{\textit{Generative Method}} & \textit{GAN-Based} & AGAA \cite{zhu2018generative} & \ding{51} & \textit{Region Annot}\\
         & \textit{VAE-Based} & AREES \cite{liu2022zero} & \ding{51} & \ding{55}\\
         \cmidrule{2-5}
         & \textit{Direct Synthesize} & Composer \cite{huynh2020compositional} & \ding{55} & \ding{55}\\
        \midrule
        \textit{Others} & \textit{Attribute Selection} & MCZSL \cite{akata2016multi} & \ding{51} & \textit{Region Annot; Online Media}\\
        \bottomrule     
    \end{tabular}}
    \centering
    \caption{The categorization of representative non-attention fine-grained zero-shot learning methods.}
    \label{tab:non-attention}
\end{table*}

\subsubsection{Local Coordination}
The motivation of local coordination is to directly generate a set of coordinates to reveal the most meaningful visual regions/parts, which can be expressed as:
\begin{equation}
    Z = [z_h, z_w, z_l] = \mathcal{F}(x),
\end{equation}
where $x$ and $\mathcal{F}$ are the visual feature and learnable network. $Z$ is the window, $z_h, z_w$ denote the coordinates, and $z_l$ denotes the length of the region. 
For example, LDF~\cite{li2018discriminative} employs a network called ZoomNet. After obtaining the coordinates of the key region, ZoomNet further zooms it to attract the attention of the training network. Differently, SGMA~\cite{zhu2019semantic} takes the attention masks as the input to get the coordinates of multiple regions and then crops the original image afterward. The cropped patches are used to assist in the network judgment.

\subsubsection{Score Function}
Attribute and visual attentions mostly adopt the strategy of independent operations, i.e., attribute and visual features are not involved in the attention computation simultaneously. 
Cross attention remedies this issue with the motivation of obtaining a more detailed attention map by densely detecting visual and attribute correlations. The score function is one of the main directions, whose gist is to compute one-to-one similarity scores between regional visual features and subattribute vectors. 
Suppose $x\in \mathbb{R}^{C\times r}$ denotes the visual feature with $r=H\times W$ regions. Let $a\in \mathbb{R}^{d\times m}$ denote the attribute vector, where $m$ is the number of attributes and $d$ is the vector dimension. Then, the similarity matrix can be expressed as $\phi(a)^{\mathsf{T}}x$, where $\phi$ denotes the mapping function to ensure that the visual and attribute vectors are in the same dimension space. The attention map can then be represented as:
\begin{equation}
    S=\frac{\exp(\phi(a)^{\mathsf{T}}x)}{\sum^r\exp(\phi(a)^{\mathsf{T}}x)},
\end{equation}
where $S\in \mathbb{R}^{m\times r}$ and $\phi(a)^{\mathsf{T}}x$ measures the degree of correlation between subattributes and each regional feature. $S$ represents the weighted matrix to suppress the influence of those regions with lower scores. 

Among this area, GEM~\cite{liu2021goal} uses the $S$ directly for the downstream task and prompts the model to focus on the specific regions under the supervision of gaze annotations. DAZLE~\cite{huynh2020fine_cvpr20}, on the other hand, multiplies $\phi(a)^{\mathsf{T}}x$ and $S$ and then applies the result to the final prediction. Derived from DAZLE, MSDN~\cite{chen2022msdn} proposes a bidirectional attention network that can further calibrate the visual and semantic domain bias. 

\subsubsection{Self Attention}
As one of the key components in Transformer~\cite{vaswani2017attention}, self attention has been extended to a wide range of areas in recent years due to its powerful ability to capture contextual dependencies~\cite{chen2023duet}. Suppose that we have \textit{Query}, \textit{Key}, and \textit{Value} denoted by $Q$, $K$, and $V$. A universal representation of self attention can be expressed as:
\begin{equation}
\mathrm{Output} = \frac{QK^{\mathsf{T}}\tau}{\sum QK^{\mathsf{T}}\tau}V,
\end{equation}
where $\tau$ is a scaling constant. 
The most critical issue in applying self attention to the FZSL task is how to design its $Q$, $K$, and $V$ based on available resources, i.e., \textit{how should the visual feature and attribute vector be treated}?
%

Several methods have been proposed to answer it. For example, TransZero~\cite{chen2022transzero} sets them all as visual features transformed by three different linear networks in the encoder, and later in the decoder as \{\textit{attribute}, \textit{visual}, \textit{visual}\}. Differently, I2DFormer~\cite{naeem2022i2dformer} adpots \{\textit{visual}, \textit{attribute}, \textit{attribute}\} as $\{Q, K, V\}$, respectively, while PSVMA~\cite{liu2023progressive} uses \{\textit{attribute}, \textit{visual}, \textit{visual}\}. In contrast, HRT~\cite{cheng2023hybrid} takes a distinct configuration of \{\textit{visual}, \textit{attribute}, \textit{class embedding}\} in the decoder.

\begin{table*}[htbp]
    \renewcommand{\arraystretch}{0.8}
    \centering
    \resizebox{\textwidth}{!}{
    \begin{tabular}{l c c c c c c c c }
        \toprule
        \textbf{\large{Name}} & \textbf{\large{Acronym}} & \textbf{\large{Granularity}} & \textbf{\large{\#Images}} & \textbf{\large{Categories}} & \textbf{\large{\#Categories}} & \textbf{\large{Seen/Unseen}} & \textbf{\large{Attribute}} & \textbf{\large{\#Attribute}}\\
        \midrule
        Caltech-UCSD-Birds$^{[1]}$ & CUB & \textit{Fine} & 11,788 & \textit{Birds} & 200 & 150/50 & \textit{Word Description} & 312 \\
        Oxford Flowers$^{[2]}$ & FLO & \textit{Fine} & 8,189 & \textit{Flowers} & 102 & 82/20 & \textit{Class Embedding} & - \\ 
        SUN Attribute$^{[3]}$ & SUN & \textit{Fine} & 14,340 & \textit{Scenes} & 717 &  645/72 & \textit{Word Description} & 102 \\
        NABirds$^{[4]}$ & - & \textit{Fine} & 48,562 & \textit{Birds} & 404$^\dagger$ & 323/81 & - & - \\
        DeepFashion$^{[5]}$ & - & \textit{Fine} & 289,222 & \textit{Clothes} & 46 & 36/10 & \textit{Word Description} & 1000\\
        Animals with Attributes$^{[6]}$ & AWA & \textit{Coarse} & 30,475 & \textit{Animals} & 50 & 40/10 & \textit{Word Description} & 85\\
        Animals with Attributes(2)$^{[7]}$ & AWA2 & \textit{Coarse} & 37,322 & \textit{Animals} & 50 & 40/10 & \textit{Word Description} & 85\\
        Attribute Pascal and Yahoo$^{[8]}$ & APY & \textit{Coarse} & 15,339 & \textit{Objects} & 32 & 20/12 & \textit{Word Description} & 64 \\
        \bottomrule

        \multicolumn{9}{p{22cm}}{\footnotesize{
        \textit{In Table Ref.}: $^{[1]}$\cite{welinder2010caltech}, $^{[2]}$\cite{nilsback2008automated}, $^{[3]}$\cite{patterson2012sun}, $^{[4]}$\cite{van2015building}, $^{[5]}$\cite{liu2016deepfashion}, $^{[6]}$\cite{lampert2013attribute_pami13},$^{[7]}$\cite{xian2018zero},$^{[8]}$\cite{farhadi2009describing}.}} \\

        \multicolumn{9}{l}{\textit{\footnotesize{Symbol Interpretation:}} \ding{172} \textit{\footnotesize{$\dagger$: Compression to fit the setting of zero-shot learning}}.} \\
    \end{tabular}}
    \centering
    \caption{A list of commonly used benchmark datasets.}
    \label{tab:datasets}
\end{table*}

\begin{table*}[htbp]
    \renewcommand{\arraystretch}{0.7}
    \centering
    \resizebox{\textwidth}{!}{
    \begin{tabular}{l l l c l l l }
        \toprule
        \textbf{\large{Method}} & \textbf{\large{Venue}} & \textbf{\large{Backbone}} & \textbf{\large{FT}} & \textbf{\large{Resolution}} & \textbf{\large{Datasets}} & \textbf{\large{Code}}\\
        \midrule
        \multicolumn{7}{c}{\textit{Attention-Based}} \\
        \midrule
         LDF$^{[1]}$ & CVPR~$'{18}$  & GNet, VGG19 & \ding{51} & $224\times 224$ & CUB, AWA & \href{https://github.com/zbxzc35/Zero_shot_learning_using_LDF_tensorflow}{github.com/zbxzc35}\\
        LFGAA$^{[2]}$ & ICCV~$'{19}$ & GNet, R101, V19 & \ding{51} & $224\times 224$ & CUB, SUN, AWA2 & \href{https://github.com/ZJULearning/AttentionZSL}{github.com/ZJULearn} \\
        AREN$^{[3]}$ & CVPR~$'{19}$  & ResNet101& \ding{51} & $224\times 224$ & CUB, SUN, AWA2, APY & \href{https://github.com/gsx0/Attentive-Region-Embedding-Network-for-Zero-shot-Learning}{github.com/gsx0}\\
        SGMA$^{[4]}$ & NeurIPS~$'{19}$  & VGG19 & \ding{51} & $448\times 448$ & CUB, FLO, AWA & \href{https://github.com/wuhuicumt/LearningWhereToLook/tree/master}{github.com/wuhuicum}\\
        RGEN$^{[5]}$ & ECCV~$'{20}$  & ResNet101& \ding{51} & $224\times 224$ & CUB, SUN, AWA2, APY & -\\
        DAZLE$^{[6]}$ & CVPR~$'{20}$  & ResNet101 & \ding{55} & $224\times 224$ & CUB, SUN, DeepFashion, AWA2 & \href{https://github.com/hbdat/cvpr20_DAZLE}{github.com/hbdat}\\
        RSAN$^{[7]}$ & CIKM~$'{21}$  & ResNet101 & - & $448\times 448$ & CUB, SUN, AWA2 & -\\
        GEM$^{[8]}$ & CVPR~$'{21}$  & ResNet101& \ding{51} & $448\times 448$ & CUB, SUN, AWA2 & \href{https://github.com/osierboy/GEM-ZSL}{github.com/osierboy}\\
        I2DFormer$^{[9]}$ & NeurIPS~$'{22}$ & ViT-B & \ding{51} & $224\times 224$ & CUB, FLO, AWA2 & \href{https://github.com/ferjad/I2DFormer}{github.com/ferjad}\\
        MSDN$^{[10]}$ & CVPR~$'{22}$  & ResNet101 & \ding{55} & $448\times 448$ & CUB, SUN, AWA2 & \href{https://github.com/shiming-chen/MSDN}{github.com/shiming}\\
        TransZero$^{[11]}$ & AAAI~$'{22}$ & ResNet101 & \ding{55} & $448\times 448$ & CUB, SUN, AWA2 & \href{https://github.com/shiming-chen/TransZero}{github.com/shiming}\\
        DUET$^{[12]}$ & AAAI~$'{23}$ & ViT-B & \ding{51} & $224\times 224$ & CUB, SUN, AWA2 & \href{https://github.com/zjukg/DUET}{github.com/zjukg}\\
        PSVMA$^{[13]}$ & CVPR~$'{23}$ & ViT-B& \ding{51} & $224\times 224$ & CUB, SUN, AWA2 & \href{https://github.com/ManLiuCoder/PSVMA}{github.com/ManLiu}\\
        \midrule
        \multicolumn{7}{c}{\textit{Prototype Learning}} \\
        \midrule
        APN$^{[14]}$ & NeurIPS~$'{20}$ & ResNet101 & \ding{51} & $224\times 224$ & CUB, SUN, AWA2 & \href{https://github.com/wenjiaXu/APN-ZSL/tree/master}{github.com/wenjiaXu}\\
        DPPN$^{[15]}$ & NeurIPS~$'{21}$ & ResNet101& \ding{51} & $448\times 448$ & CUB, SUN, AWA2, APY & \href{https://github.com/Roxanne-Wang/DPPN-GZSL}{github.com/Roxanne}\\
        DPDN$^{[16]}$ & MM~$'{22}$ & ResNet101 & \ding{55} & $448\times 448$ & CUB, SUN, AWA2 & -\\
        CoAR-ZSL$^{[17]}$ & TNNLS~$'{23}$ & ResNet101,ViT-L & \ding{51} & $448\times 448^*$ & CUB, SUN, AWA2 & \href{https://github.com/dyabel/CoAR-ZSL}{github.com/dyabel}\\
         \midrule
        \multicolumn{7}{c}{\textit{Data Manipulation}} \\
        \midrule
        LH2B$^{[18]}$ & CVPR~$'{17}$ & VGG16 & \ding{55} & - & CUB, NABirds & \href{https://github.com/EthanZhu90/ZSL_PP_CVPR17}{github.com/EthanZhu}\\
        S2GA$^{[19]}$ & NeurIPS~$'{18}$ & VGG16 & \ding{55} & - & CUB, NABirds & \href{https://github.com/ylytju/sga/tree/master}{github.com/ylytju}\\ 
        SR2E$^{[20]}$ & AAAI~$'{21}$ & ResNet101 & - & $448\times 448$ & CUB, SUN, AWA2, APY & -\\
        VGSE-SMO$^{[21]}$ & CVPR~$'{22}$ & ResNet50& - & - & CUB, SUN, AWA2 & \href{ https://github.com/wenjiaXu/VGSE}{github.com/wenjiaXu}\\
        \midrule
        \multicolumn{7}{c}{\textit{Graph Modeling}} \\
        \midrule
        APNet$^{[22]}$ & AAAI~$'{20}$ & ResNet101 & \ding{55} & - & CUB, SUN, AWA, AWA2, APY & -\\
        GNDAN$^{[23]}$ & TNNLS~$'{22}$ &  ResNet101 & \ding{55} & $448\times 448$ & CUB, SUN, AWA2 & \href{https://github.com/shiming-chen/GNDAN}{github.com/shiming}\\
        GKU$^{[24]}$ & AAAI~$'{23}$ & ResNet34 & - & - & CUB, NABirds & -\\
        EOPA$^{[25]}$ & TPAMI~$'{23}$ & ANet, ResNet50 & \ding{51} & - & CUB, SUN, FLO, AWA2 & -\\
        \midrule
        \multicolumn{7}{c}{\textit{Generative Method}} \\
        \midrule
        AGAA$^{[26]}$ & CVPR~$'{18}$ & VGG16 & \ding{55} & $224\times 224$ & CUB, NABirds & \href{https://github.com/EthanZhu90/ZSL_GAN}{github.com/EthanZhu}\\
        Composer$^{[27]}$ & NeurIPS~$'{20}$ & ResNet101 & \ding{55} & $224\times 224$ & CUB, SUN, DeepFashion, AWA2 & \href{https://github.com/hbdat/neurIPS20_CompositionZSL}{github.com/hbdat}\\
        AREES$^{[28]}$ & TNNLS~$'{22}$ & ResNet101 & \ding{55} & $224\times 224$ & CUB, SUN, AWA, AWA2, APY & -\\
        \midrule
        \multicolumn{7}{c}{\textit{Others}} \\
        \midrule
        MCZSL$^{[29]}$ & CVPR~$'{16}$ & VGG16 & \ding{55} & $224\times 224$ & CUB & -\\
        \bottomrule

        \multicolumn{7}{p{20cm}}{\footnotesize{
        \textit{In Table Ref.}: $^{[1]}$\cite{li2018discriminative}, 
        $^{[2]}$\cite{liu2019attribute}, 
        $^{[3]}$\cite{xie2019attentive}, 
        $^{[4]}$\cite{zhu2019semantic}, 
        $^{[5]}$\cite{xie2020region}, 
        $^{[6]}$\cite{huynh2020fine_cvpr20}, 
        $^{[7]}$\cite{wang2021region},
        $^{[8]}$\cite{liu2021goal}, 
        $^{[9]}$\cite{naeem2022i2dformer}, 
        $^{[10]}$\cite{chen2022msdn}, 
        $^{[11]}$\cite{chen2022transzero}, 
        $^{[12]}$\cite{chen2023duet},
        $^{[13]}$\cite{liu2023progressive},
        $^{[14]}$\cite{xu2020attribute}, 
        $^{[15]}$\cite{wang2021dual}, 
        $^{[16]}$\cite{ge2022dual}, 
        $^{[17]}$\cite{du2023boosting},
        $^{[18]}$\cite{elhoseiny2017link},
        $^{[19]}$\cite{ji2018stacked_nips18},
        $^{[20]}$\cite{ge2021semantic},
        $^{[21]}$\cite{xu2022vgse}, 
        $^{[22]}$\cite{liu2020attribute}, 
        $^{[23]}$\cite{chen2022gndan},
        $^{[24]}$\cite{guo2023graph}, 
        $^{[25]}$\cite{chen2023explanatory},
        $^{[26]}$\cite{zhu2018generative},
        $^{[27]}$\cite{huynh2020compositional}, 
        $^{[28]}$\cite{liu2022zero},
        $^{[29]}$\cite{akata2016multi}.}} \\
        
        \textit{\footnotesize{Symbol Interpretation:}} & \multicolumn{6}{l}{\ding{172} \textit{\footnotesize{GNet: GoogLeNet; R101: ResNet101; V19: VGG19; ANet: AlexNet}}.} \\
        & \multicolumn{6}{l}{\ding{173} $*$\textit{\footnotesize{: Both $224\times 224$ and $448\times 448$ resolutions are used}}.} \\
    \end{tabular}}
    \centering
    \caption{A library of fine-grained zero-shot learning methods.}
    \label{tab:library}
\end{table*}

\subsection{Non-Attention Methods}
As demonstrated in Table~\ref{tab:non-attention}, we categorize representative non-attention methods of FZSL and further tag the secondary areas according to their specific implementation strategies.

\subsubsection{Prototype Learning}
The gist of prototype learning is to assign an exemplar
to each subattribute to alleviate the issue of domain bias between global visual features and class semantic embeddings. Depending on the way prototype features are learned, methods in such areas can be categorized as Prototype-Independent~\cite{xu2020attribute,cheng2023discriminative,du2023boosting} and Prototype-Symbiotic~\cite{wang2021dual,ge2022dual,guo2023group}. 
Specifically, \textbf{Prototype-Independent} indicates that the learning processes of prototype features and sample features are independent of each other. For example, APN~\cite{xu2020attribute} utilizes regression loss to drive the model to learn prototype-related regional features while using decorrelation loss to constrain the independence of each prototype. 
In contrast, \textbf{Prototype-Symbiotic} defines that the sample features will participate in the update of the prototype features in a joint manner. For example, DPPN~\cite{wang2021dual} designs a parametric network to iteratively optimize the prototype pool.

\subsubsection{Data Manipulation}
Similar to the attention mechanism that focuses on local regions, data manipulation adopts other strategies to extract key local information from samples. Methods in such areas include Patch Clustering, Detector-Based, and Image Crop. 
Specifically, the \textbf{Patch Clustering}, e.g., VGSE-SMO~\cite{xu2022vgse}, utilizes an unsupervised segmentation algorithm to slice the image into several patches, after which the corresponding attribute semantics are learned for the patch clusters. 
Differently, \textbf{Detector-Based} methods resort to detection networks to pinpoint critical regions~\cite{elhoseiny2017link,ji2018stacked_nips18}. However, these approaches require the support of region or key point annotations. 
Last, the goal of \textbf{Image Crop} methods is to find the optimal way for sample cropping. For example, SR2E~\cite{ge2021semantic} instantiates this goal as a serialized search task in the action space, while ERPCNet~\cite{li2022entropy} incorporates the idea of reinforcement learning, which guides the model to discover the most valuable parts by setting reasonable reward targets.

\subsubsection{Graph Modeling}
Graph Convolutional Networks (GCNs)~\cite{kipf2016semi} have received widespread attention in recent years due to its superior structural information aggregation capability and ingenious unstructured data processing patterns. 
Suppose $W^{(l)}$ denotes the parameters of the $l$-th layer of GCNs, the output of the $(l+1)$-layer can be expressed as:
\begin{equation}
    H^{(l+1)}=\sigma(\tilde{D}^{-\frac{1}{2}}\tilde{A}\tilde{D}^{-\frac{1}{2}}H^{(l)}W^{(l)}),
\end{equation}
where $A$ is the adjacent matrix and $\tilde{D}$ denotes its degree matrix. $\sigma$ denotes the activation function, and $H^{(l)}$ is the output of the $l$-layer of GCNs. In FZSL, region features are naturally available as nodes for the graph. Inspired by it, the \textbf{Visual Enhancement} methods aim to aggregate local information to improve feature discrimination. For example, some studies~\cite{hu2022region,chen2022gndan} adaptively aggregate features by similarity metrics, while GKU~\cite{guo2023graph} performs graph modeling on key nodes under the supervision of region annotations. 
Differently, APNet~\cite{liu2020attribute} applies graph modeling for \textbf{Attribute Enhancement}, such a group of methods is motivated by mining the intrinsic relationships of attribute descriptions to obtain more discriminative attribute representations. 
In contrast, EOPA~\cite{chen2023explanatory} devotes to \textbf{Region Search}, which automates the search of region features corresponding to attributes by constructing a multi-granularity hierarchical graph.

\subsubsection{Generative Method}
Simulating unseen class samples with the help of Generative Adversarial Networks~(\textbf{GANs}) or Variational Autoencoders~(\textbf{VAEs}) is another important direction in FZSL. Conventional generative methods learn relationships between global features and class-wise attributes, neglecting fine-grained knowledge~\cite{li2023vs_ijcai23}. 
To resolve the issue, AGAA~\cite{zhu2018generative} leverages a detection network to extract and combine multiple critical region features as real samples, which improves the generation quality. AREES~\cite{liu2022zero} utilizes the attention mechanism to guide the model to focus on partial regions, thus enhancing the generation effect. 
In addition, Composer~\cite{huynh2020compositional} proposes a~\textbf{Direct Synthesize} scheme, which first employs the attention approach to locate the relevant regional features of attributes and then synthesizes the samples of unseen classes by combining these features. 

\subsubsection{Attribute Selection}
\noindent{MCZSL}~\cite{akata2016multi} argues that manually annotated fine-grained attributes are expensive and time-consuming, and therefore proposes to search textual descriptions of categories from online media, such as Wikimedia. Due to the poor quality of attributes obtained in this way, it devises multiple methods to filter the noise. 

\section{Library}
\label{library}
We further systematically summarize the common benchmarks in FZSL, including widely used datasets, representative models, implementations, and some more details in a nutshell, and provide an FZSL repository to enrich the community resources. It is expected that such resources can assist researchers with better access to existing approaches and faster implementation of FZSL research. The open library is publicly accessible via \url{https://github.com/eigenailab/Awesome-Fine-Grained-Zero-Shot-Learning}.

\subsection{Datasets}
Table~\ref{tab:datasets} shows the commonly used benchmark datasets, including $5$ fine-grained and $3$ coarse-grained datasets. We list the detailed configuration information, including the total number of samples, sample types, the total number of categories, the split of seen/unseen categories, attribute types, and dimensions. 
Within the table, \textit{Word Description} denotes professional-level annotations, e.g., CUB contains $312$ terms describing birds such as \textit{\{has bill shape::hooked, has wing color::red, has breast pattern::solid\}}. \textit{Class Embedding} denotes the semantic feature obtained with the category name. In fact, FLO also contains fine-grained text annotations, i.e., $10$ sentences per image. NABirds has $1011$ classes, which are compressed to $404$ classes due to category overlap. NAbirds has no attribute annotations, but has region annotations.

\subsection{Details}
We collect relevant details from the literature on fine-grained zero-shot learning to provide a more comprehensive reference for the model implementation. 
As demonstrated in Table~\ref{tab:library}, we elaborate on the basic experimental setup of representative methods. Specifically, as to the \textbf{Backbone and FT} (i.e., Finetune), we list the backbone networks used as the feature extractor (excluding the downstream classifier), and FT indicates whether the feature extractor is involved in training or not. The crossmark \ding{55} represents that the network pre-trained on ImageNet is used as the feature extractor, and its parameters are fixed. The \textbf{Resolution} indicates the size of input images, and \textbf{Datasets} lists the datasets evaluated in experiments. Last, \textbf{Code} attaches the links to open source codes (in any) of representative methods to facilitate access.

\section{Application}
\label{application}
With the purpose of serving open environments with restricted visual samples and the core of attribute primitive-driven research, FZSL has expanded to various applications and enlightened a range of related academic areas. 
Some representative applications include but not limited to
1)~\textbf{Low-Shot Object Recognition}: FZSL methods are naturally adapted to other variants of ZSL, such as Transductive ZSL~\cite{yao2021attribute}, Compositional ZSL~\cite{panda2024compositional}, and Multi-Label ZSL~\cite{huynh2020shared}. Meanwhile, the ideology of FZSL fits seamlessly into a variety of data-constrained scenarios, such as semi-supervised learning~\cite{huynh2020interactive}, few-shot learning~\cite{wu2023hierarchical}, and transfer learning~\cite{liu2024fine}. 
2)~\textbf{Scene Understanding}: Object detection and semantic segmentation are two critical and complex scene understanding tasks whose performance benefits from massive and meticulous scene annotations. To release the heavy annotation pressure as well as adapt to the requirement of out-of-distribution (OOD) detection, the research that combines FZSL and scene understanding emerges as a promising direction and has received increasing attention~\cite{bansal2018zero,he2023primitive}. 
3)~\textbf{Open Environment Application}: In addition to the field of natural image recognition, FZSL has also driven the application and development of a series of special tasks to accommodate the open environment. To name a few, medical~\cite{mahapatra2022self} and remote sensing~\cite{sumbul2017fine}, video classification~\cite{hong2023fine}, and action recognition~\cite{chen2021elaborative}. 
4)~\textbf{Model Robustness}: More than just the performance, the robustness of models in FZSL has recently attracted the interest of increasing researchers to expose weaknesses by applying adversarial learning~\cite{shafiee2022zero}.

\section{Challenges and Opportunities}
\label{c_and_o}

In this paper, we comb the studies of the last decade on integrating fine-grained analysis into ZSL and exhibit their core contributions in an organized manner. From mining local visual features and capturing fine-grained relations to reconstructing attribute spaces, FZSL researchers have provided a large number of promising solutions around the three realms of analysis, including visual, attribute, and mapping function. However, several limitations imply the imperfect development of FZSL as well as the direction of future opportunities.

\subsubsection{Annotation Cost and Quality}
Fine-grained attribute learning requires extensive refined annotations. However, the attribute-level annotations are time- and labor-intensive compared to class-level labeling. Worse still, once FZSL settles into concrete real-world scenarios, such as industrial inspection or medical pathology, the expert knowledge can be a bottleneck, which further raises the labor cost. In addition, attribute engineering is a complex crossover field. Even attributes annotated by experienced experts do not guarantee benefits for deep learning, which implies that high-quality attribute annotations require professionals with dual knowledge of both specific domains and deep learning. Despite some studies attempting to make breakthroughs in the field of automated annotation~\cite{akata2016multi}, it is clear that there is still a long way to go.

\subsubsection{Deployment Cost}
Compared to class-wise semantic modeling, FZSL typically has to process a higher density of information, which introduces a more luxurious deployment cost. Such cost is reflected in bloated network structures and high computational complexity (\textit{Note that we discuss the deployment phase, excluding the training phase}). As a result, most FZSL approaches are unfriendly to edge tasks and mini-endpoints, which have to trade off performance and memory. 
However, FZSL can be more favorable to a scenario associated with resource-constrained devices due to the low or even zero data requirements. Such a scenario can also well align with ubiquitous devices and data in real-world applications. 
Therefore, it is promising to investigate on-device-friendly algorithms.

\subsubsection{Poor Theoretical Foundation}
The development of FZSL is established on the beautiful hypothesis that deep neural networks can reason logically like humans, like inferring zebra characteristics from the color of a panda, the morphology of a horse, and the stripes of a tiger. Nevertheless, there are not many solid theories on the compatibility between human reasoning and machine inductive ability so far, leading to a lack of explainability. Meanwhile, some flaws also challenge the plausibility of the hypothesis, such as the correspondence between abstract attributes and vision. Rigorous theoretical guidance is at the helm of a field moving forward, and it is of great prospective to dive into the mysterious black box in the future.

\section*{Acknowledgments}
This research was supported by funding from the Hong Kong RGC General Research Fund (GRF-No.~152211/23E), the National Natural Science Foundation of China (NSFC-No.~62102327), and the Hong Kong Polytechnic University Internal Fund (No.~P0043932,~P0038289,~and~P0043038).

\bibliographystyle{named}
\bibliography{ijcai24}

\end{document}